\documentclass[letterpaper, preprint, paper,11pt]{AAS}	

\usepackage{bm}
\usepackage{amsmath}
\usepackage{subfigure}
\usepackage[colorlinks=true, pdfstartview=FitV, linkcolor=black, citecolor= black, urlcolor= black]{hyperref}
\usepackage{overcite}
\usepackage{footnpag}			      	

\PaperNumber{22-775}

\begin{document}

\title{Satellite Detection in Unresolved Space Imagery for Space Domain Awareness Using Neural Networks}

\author{Jarred Jordan\thanks{Undergraduate, Space Technologies Lab, Embry-Riddle Aeronautical University, 1 Aerospace Blvd., Daytona Beach, FL, 32114.},
Daniel Posada\thanks{Ph.D. Candidate, Space Technologies Lab, Embry-Riddle Aeronautical University, 1 Aerospace Blvd., Daytona Beach, FL, 32114.},
David Zuehlke\footnotemark[2],
Angelica Radulovic\footnotemark[1],
Aryslan Malik\thanks{Visiting Professor, Space Technologies Lab, Aerospace Engineering Department, Embry-Riddle Aeronautical University, 1 Aerospace Blvd., Daytona Beach, FL, 32114.},
\ and Troy Henderson\thanks{Associate Professor, Space Technologies Lab, Aerospace Engineering Department, Embry-Riddle Aeronautical University, 1 Aerospace Blvd., Daytona Beach, FL, 32114.}
}

\maketitle{}

\begin{abstract}
This work utilizes a MobileNetV2 Convolutional Neural Network (CNN) for fast, mobile detection of satellites, and rejection of stars, in cluttered unresolved space imagery. First, a custom database is created using imagery from a synthetic satellite image program and labeled with bounding boxes over satellites for ``satellite-positive'' images. The CNN is then trained on this database and the inference is validated by checking the accuracy of the model on an external dataset constructed of real telescope imagery. In doing so, the trained CNN provides a method of rapid satellite identification for subsequent utilization in ground-based orbit estimation.
\end{abstract}

\section{Introduction}
Classification and detection of satellites in space imagery is important for various use cases such as safety, reconnaissance, contingency planning, space, and debris removal. With the increasing number of satellites in orbit especially due to new super constellations such as SpaceX's StarLink, and OneWeb, rapid and efficient methods of processing space situational awareness (SSA) observations must be developed. Current estimates put around 700,000 objects larger than 1cm in orbit, the vast majority consisting of uncontrolled debris.\cite{schildknechtOpticalSurveys_2007,Virtanen2016StreakDetection-fz} Various methods have been used for processing optical observations of resident space objects (RSOs). Some methods include, gross motion of objects across frames, inertial frame angular observations, and template matching.\cite{Brad_Sease2015-ge,Zuehlke2020-em,Zuehlke2021-uv,Virtanen2016-we,Zuehlke2020EndtoEnd-uz,Alan_Lovell2021Processing-dm}   
The core of any Space Domain Awareness (SDA) or SSA algorithm involves finding the camera frame location of an RSO in an image and segmenting the RSO from the background star-field. The detection and classification of satellites, space stations, and/or space debris using Computer Vision (CV), Machine Learning (ML), and Artificial Intelligence (AI) algorithms has recently spurred interest in space imagery. Some methods rely on different CNN architectures and Phyiscs Informed Neural Networks (PINNs). Fletcher et. al. used a Convolutional Neural Network (CNN) to detect Geosynchronous objects in unresolved space imagery.\cite{Fletcher2019-nu} And it was shown that the algorithm outperformed the classical SExtractor object segmentation algorithm on a large database of imagery of Geostationary (GEO) satellites.\cite{Fletcher2019-nu} Kyono and Fletcher also studied resolved imagery, and partially resolved imagery to estimate the quality of optical images of satellites.\cite{Kyono2020Quality-nz} Peng proposed a method of using Gaussian processes to augment orbit determination methods by learning the offset from lower fidelity predictions to true observations, and is applicable to the orbit determination steps to be considered in the future.\cite{Peng2018_Artificial_Neural-hk,Peng2019_Gaussian_Processes_for_Improving-kn} Another study looked at the possibility of using imagery from star-trackers for SSA purposes and used a Recurrent-Convolutional Neural Network (R-CNN) to estimate RSO position and velocities.\cite{Dave_undated-gy_Machine_Learning_Implementation_RSO_OD} Small segments of each image were passed through a CNN for object detection, and then the Recurrent Neural Network (RNN) was used to detect the sequence of observations of an RSO; however, all images in the database were synthetic.\cite{Dave_undated-gy_Machine_Learning_Implementation_RSO_OD} Woodward, extended the work of Fletcher and trained a CNN for pixelwise segmentation and detection of RSOs using the same SatNet database from ref. \cite{Fletcher2019-nu,Douglas_Woodward_Celeste_Manughian-Peter_Timothy_Smith_and_Elizabeth_Davison_undated-dx} 

This work is aimed to generate a database of unresolved space imagery containing both images with and without RSOs, train a MobileNetV2 CNN using the custom database, validate the CNN on external images set, assess its performance, and finally discuss future work where it is planned to build upon this CNN framework to extract the orbital elements set for orbit determination using novel Product of Exponentials (PoE) orbital mechanics framework \cite{malik2022using}. The MobileNetV2 CNN architecture was chosen for satellite detection because it has a fewer number of parameters and is faster than most of the alternative models \cite{Sandler2019MobileNetV2IR,dust_detection_2022}. Furthermore, MobileNetV2 does not require considerable computational resources to train and infer, which makes it suitable for on-orbit training and detection. As a result, a goal for this work was training sample efficiency and performance on a limited dataset to simulate such conditions. This CNN model design can be utilized for a variety of tasks, including segmentation, but it was employed for object detection in this work.

\section{Methodology}
\subsection{Space Imagery Database}
Electro-optical (EO) imagery is a common source of data for SSA applications. The Space Surveillance Network (SSN) is the largest provider of data for orbit estimation and is tasked with keeping track of orbiting objects.\cite{Vallado2007_Fundamentals-tl} A ground-based telescope system using an 11 inch Celestron RASA telescope was used to capture the images used for the dataset in this research.\cite{Richard_Berry_and_the_Celestron_Engineering_Team2016-rt} Images of varying exposure times and tracking modes were utilized to create a diverse database for model validation and testing. Images were captured in two separate satellite-rate tracking modes. The first is where the telescope tracks at a rate sufficient to keep the RSO in the telescope field of view (FOV) forming a point-source, while stars form streaks in the image. In the second, images were taken in sidereal tracking mode wherein stars are point-sources and RSOs form streaks. For a more in depth discussion of RSO tracking modes and example images see ref. \cite{Zuehlke2019-mv} 

\begin{figure}[!htb]
    \centering
    \begin{minipage}{0.5\textwidth}
        \centering
        \includegraphics[width=0.92\textwidth]{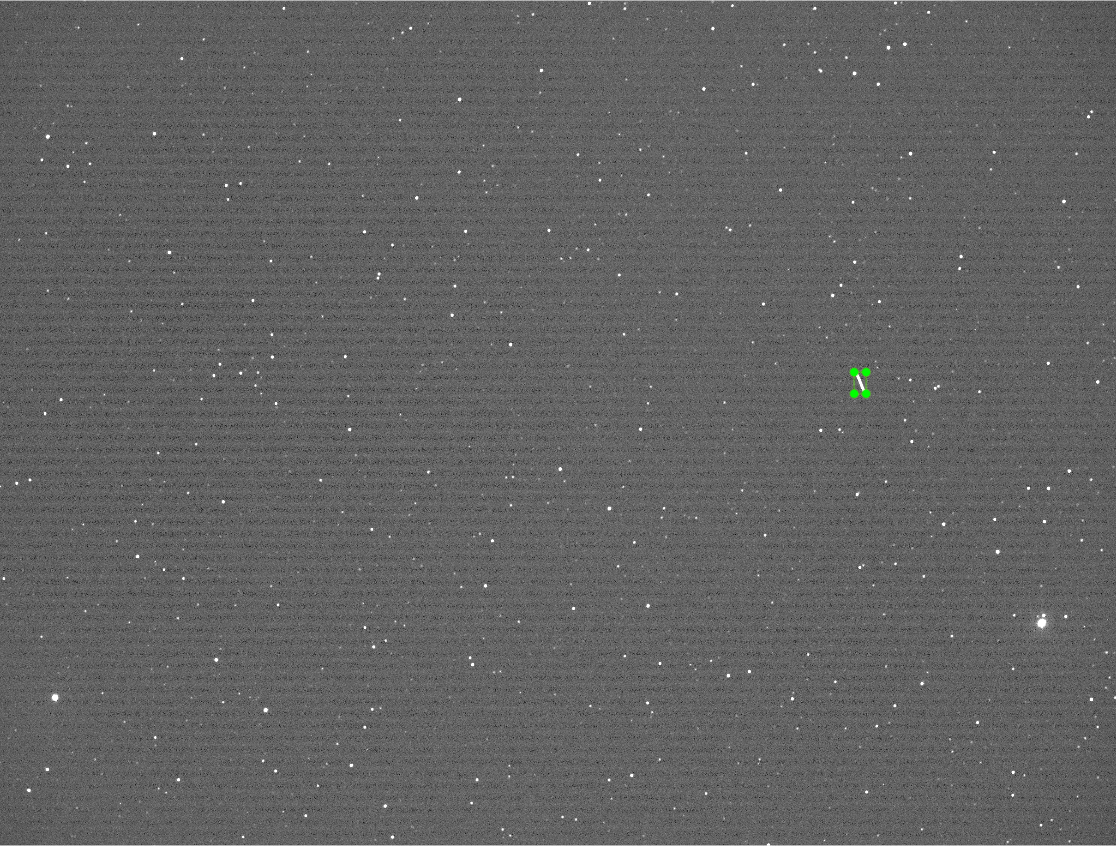}
    \end{minipage}%
    \begin{minipage}{0.5\textwidth}
        \centering
        \includegraphics[width=\textwidth]{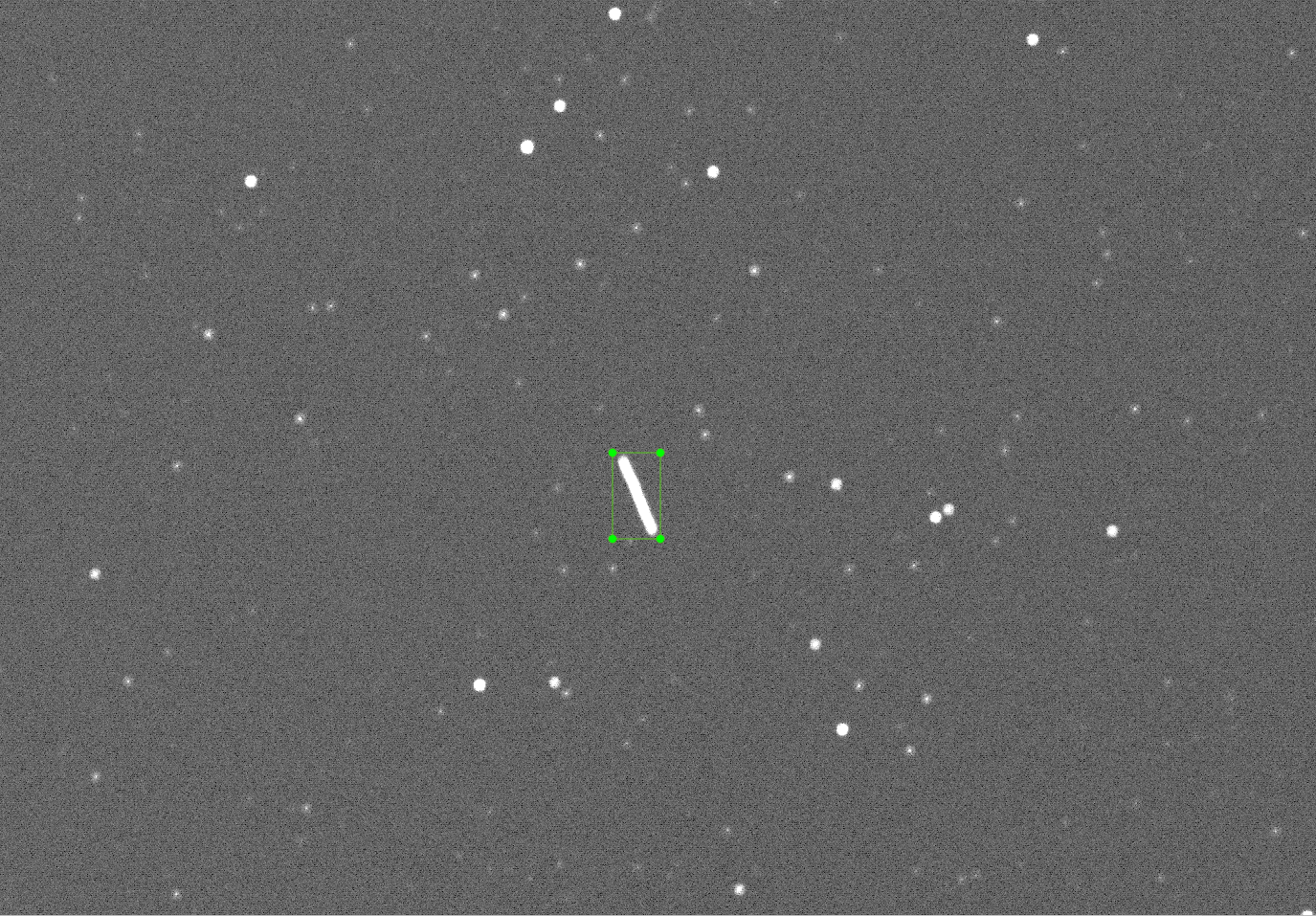}
    \end{minipage}
        \centering
    \begin{minipage}{0.5\textwidth}
        \centering
        \includegraphics[width=0.92\textwidth]{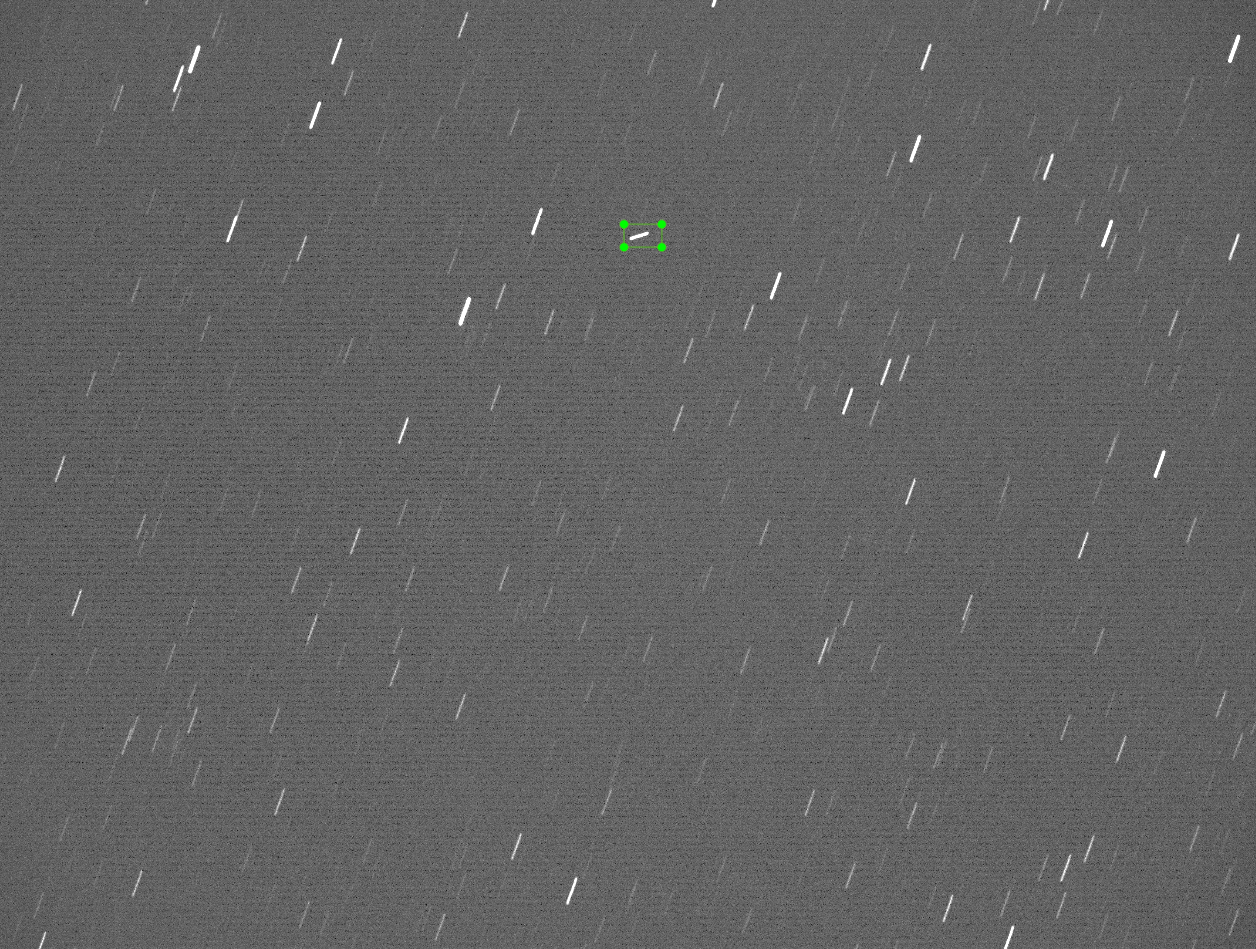}
    \end{minipage}%
    \begin{minipage}{0.5\textwidth}
        \centering
        \includegraphics[width=\textwidth]{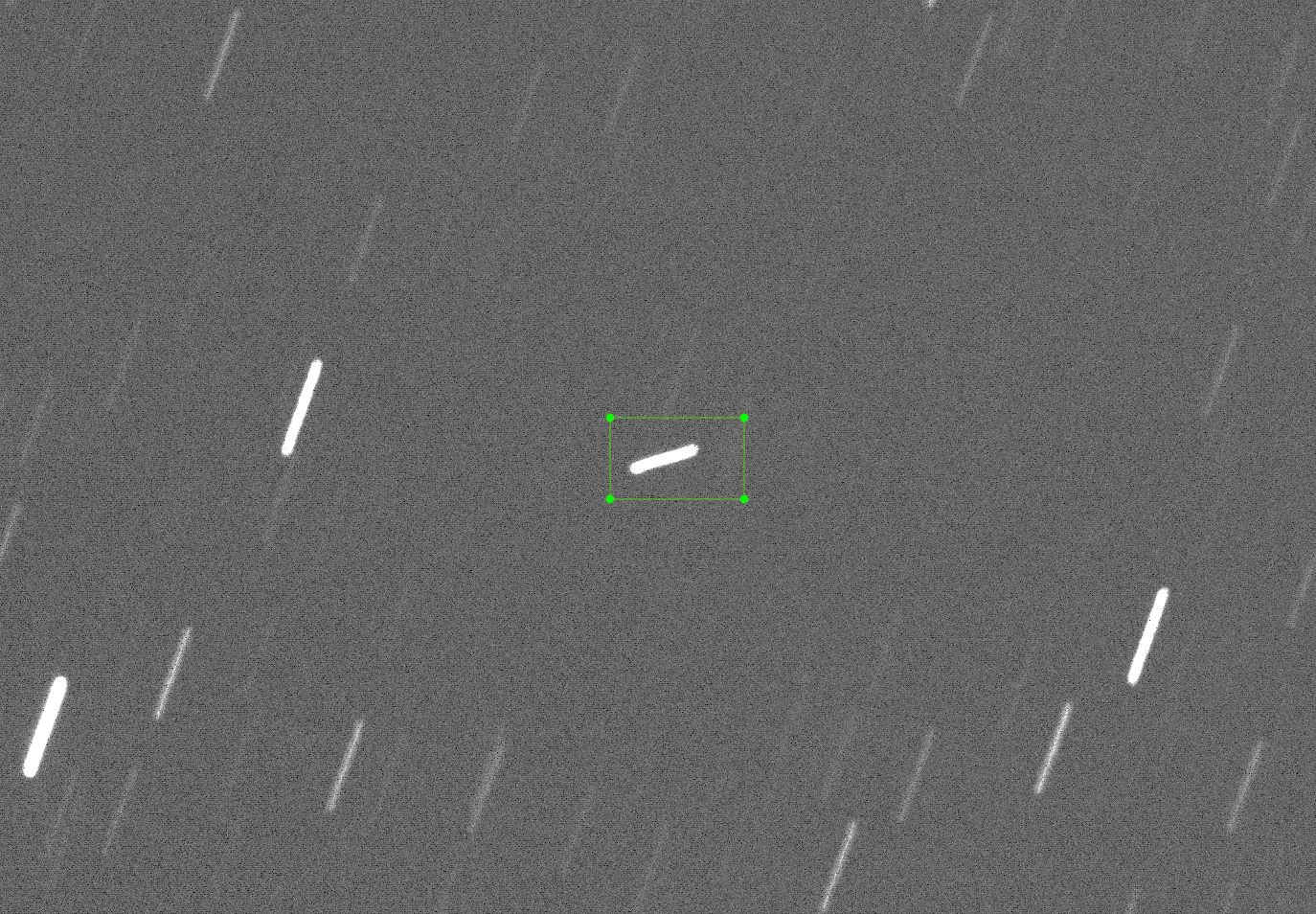}
    \end{minipage}
        \centering
    \begin{minipage}{0.5\textwidth}
        \centering
        \includegraphics[width=0.92\textwidth]{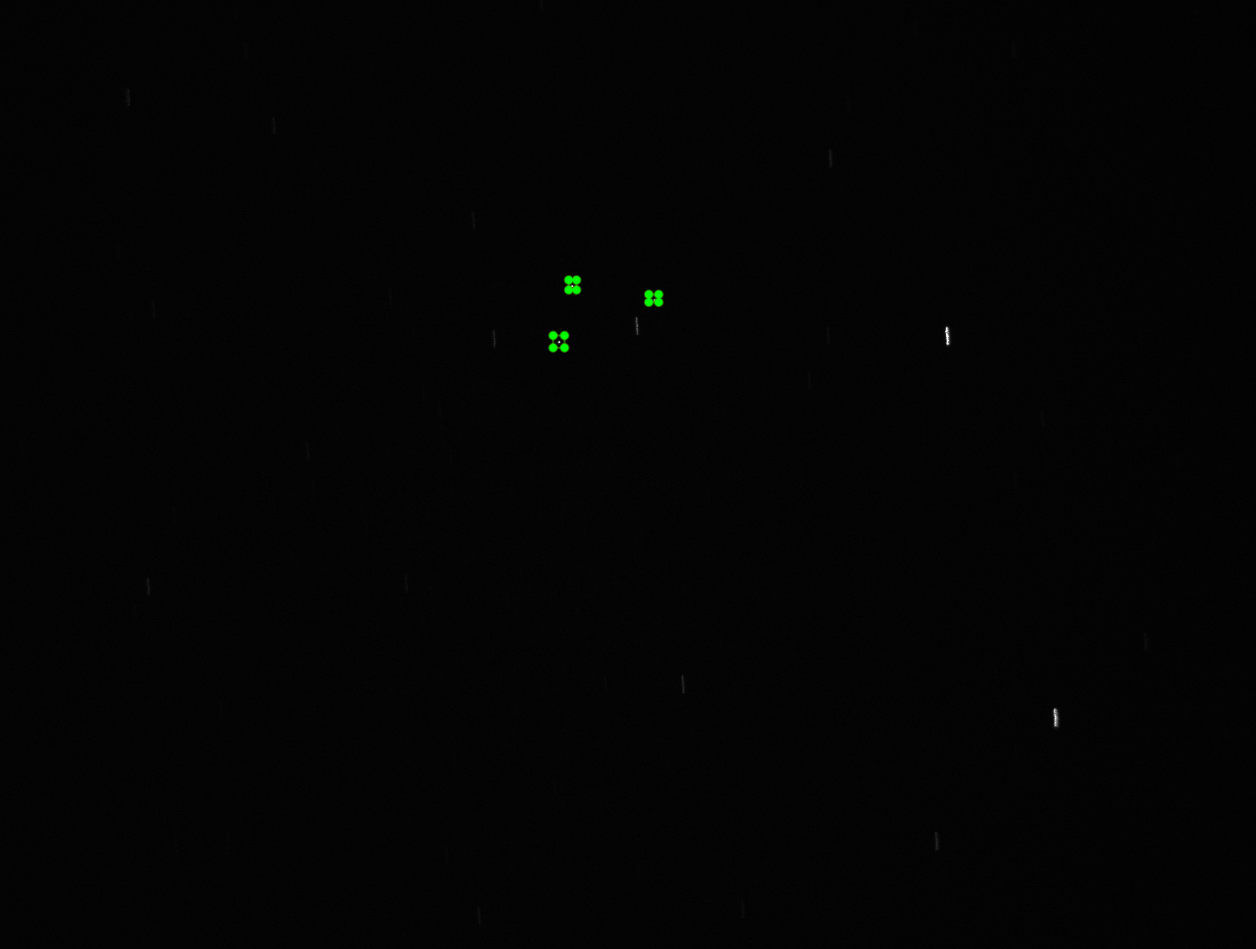}
    \end{minipage}%
    \begin{minipage}{0.5\textwidth}
        \centering
        \includegraphics[width=\textwidth]{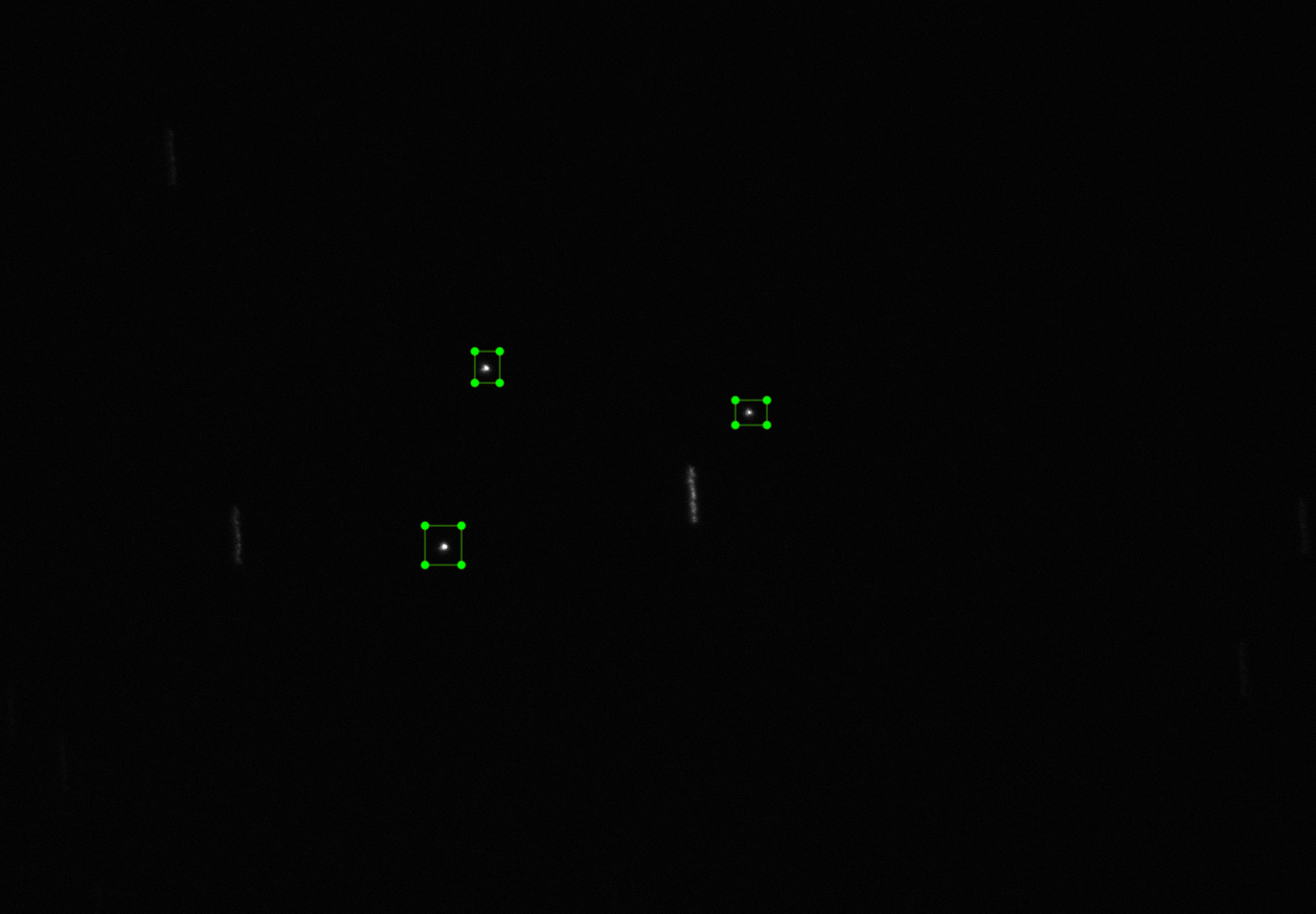}
    \end{minipage}
    \caption{RSO labeling within telescope imagery using different tracking configurations. Left: Original image. Right: Close-up of detected RSOs}
    \label{fig:dataset}
\end{figure}

In total, about 120 generated images will be labeled and used for training. Depending on network results, further labeling will be performed by trained analysts. Meanwhile, the remaining images will be saved for validation of the finished trained network. The full database of space images contains in excess of 100,000 images which vary in form from images containing no RSOs, multiple RSOs, and are portrayed in varying tracking modes. Images not used for this research can be used for future training and validation experiments. Furthermore, the database contains images of RSOs from various orbital regimes, including low-earth-orbit (LEO), medium-earth-orbit (MEO), and GEO objects. Training a neural network on a variety of these orbits will help in making the classification and detection method applicable to any SSA data. The satellites in the training images were labeled using standard bounding boxes as shown in Figure \ref{fig:dataset}. 

\subsection{Machine Learning Architecture}
To increase the accuracy of the detection, multiple images with RSOs are labeled under different tracking situations. In Figure \ref{fig:dataset} it can be seen that stars and RSOs can be either a defined ellipse (point-source) or a streak. This will be significant to make the architecture robust to different image capture settings.
The architecture is based on the state-of-the-art deep learning model MobileNetV2. This network has multiple layers that perform convolutions and bottlenecks to map the image implicit information into different dimensions. This mapping allows the model to find unique patterns that give the detection model the ability to discern between an RSO and a star.\cite{Sandler2019MobileNetV2IR, karadal2021automated} A feature of this model is the inclusion of a single shot detector (SSD) framework which presents a variety of advantages when compared to other pre-trained object detection frameworks\cite{LiuWei2016SSSM}. Moreover, SSD reinforced models are the only models currently compatible with TensorFlow Lite (TFLite) conversion requirements. This is important as these converted models are specially optimized for mobile inference and embedded systems. There are also additional improvements to resource overhead and inference latency when compared to unconverted TF models. 

\section{Model Training and Procedure}

\subsection{Hardware}
The workstation used for training the detection models contained two Intel Xeon Silver 4214 processors, 128 GB of RAM, and an NVIDIA Quadro RTX 4000. For the mobile application of the model, the tests utilized a Raspberry Pi 4 with 4 GB of system RAM and a slightly overclocked CPU which operated at 1600 MHz. No special cooling methods were needed. Additionally, the Raspberry Pi was operated at standard ambient conditions and did not utilize any external computation modules.

\subsection{Simulation Dataset}
In order to build a robust training dataset, synthetic satellite imagery was generated using SatSim\cite{cabello2022satsim}. This synthetic data generation tool allows for accurate recreation and simulation of various types of imagery with a high degree of customization including: RSO quantity, image noise, telescope tracking mode, and various other parameters. When creating the final training dataset, a setting for 50 random observations with 10 frames per observation was used to generate a wide, generalized dataset to avoid overfitting a particular neural network model during training. Since star location was not important for the models, only satellites were labeled; therefore, all models consist of a single class for detection.

As efficiency and ease of sample data acquisition are important for mobile implementations, the models were trained on a limited sample of 10 original images with an additional 5 being used for validation. While this makes preparing the model easier, there is a possible lack of sufficient variety for the model to properly learn the detection target. For a model to be robust and reliable under multiple tracking conditions and RSO visibility, training diversity is required. This was especially seen in early versions of the satellite detection model where the RSO would be completely undetected if the image orientation was slightly shifted or rotated despite using a larger set of training images at the time. In experimentation, the behavior was specifically apparent when the RSOs were streaks in the image. To fix this, the images were fed through a custom augmentation script of rotations and flips. As a result, the quantity of unique images was increased by a factor of 8 to a final training-validation ratio of 80:40. 

\begin{figure}[!h]
    \centering
        \centering
    \begin{minipage}{0.5\textwidth}
        \centering
        \includegraphics[width=0.8\textwidth]{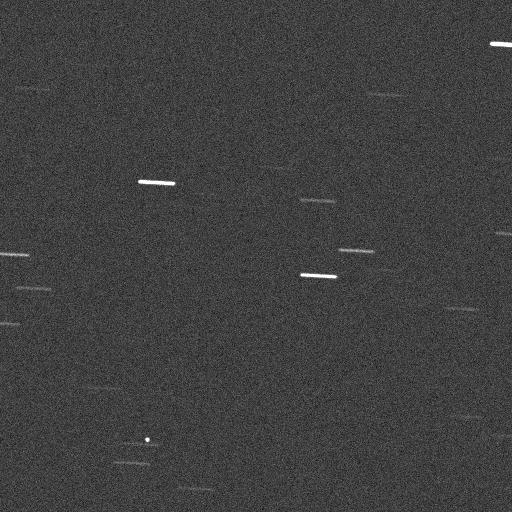}
    \end{minipage}%
    \begin{minipage}{0.5\textwidth}
        \centering
        \includegraphics[width=0.8\textwidth]{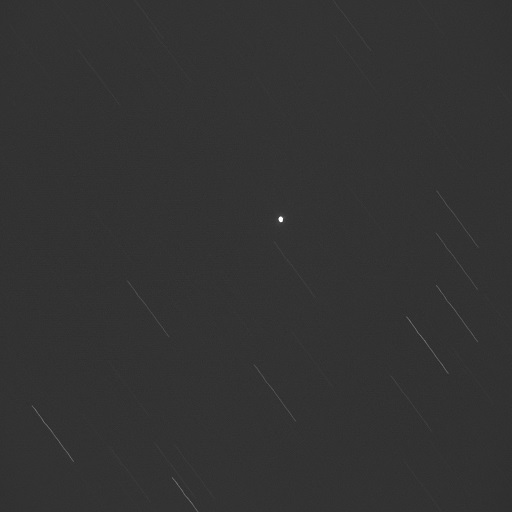}
    \end{minipage}
        \centering
    \begin{minipage}{0.5\textwidth}
        \centering
        \includegraphics[width=0.8\textwidth]{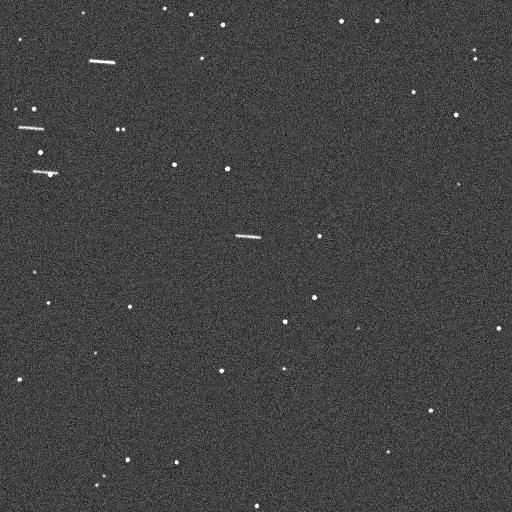}
    \end{minipage}%
    \begin{minipage}{0.5\textwidth}
        \centering
        \includegraphics[width=0.8\textwidth]{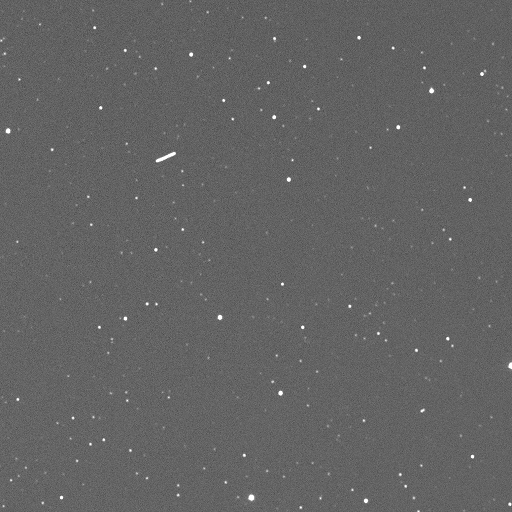}
    \end{minipage}
    \caption{Generated images uses SatSim under different tracking modes and variable RSO population (left) compared to real imagery (right)}
    \label{fig:SatSim}
\end{figure}

\subsection{Training Workflow}
\begin{figure}[h!]
    \centering
    \begin{minipage}{0.9\textwidth}
        \centering
        \includegraphics[width=.9\textwidth]{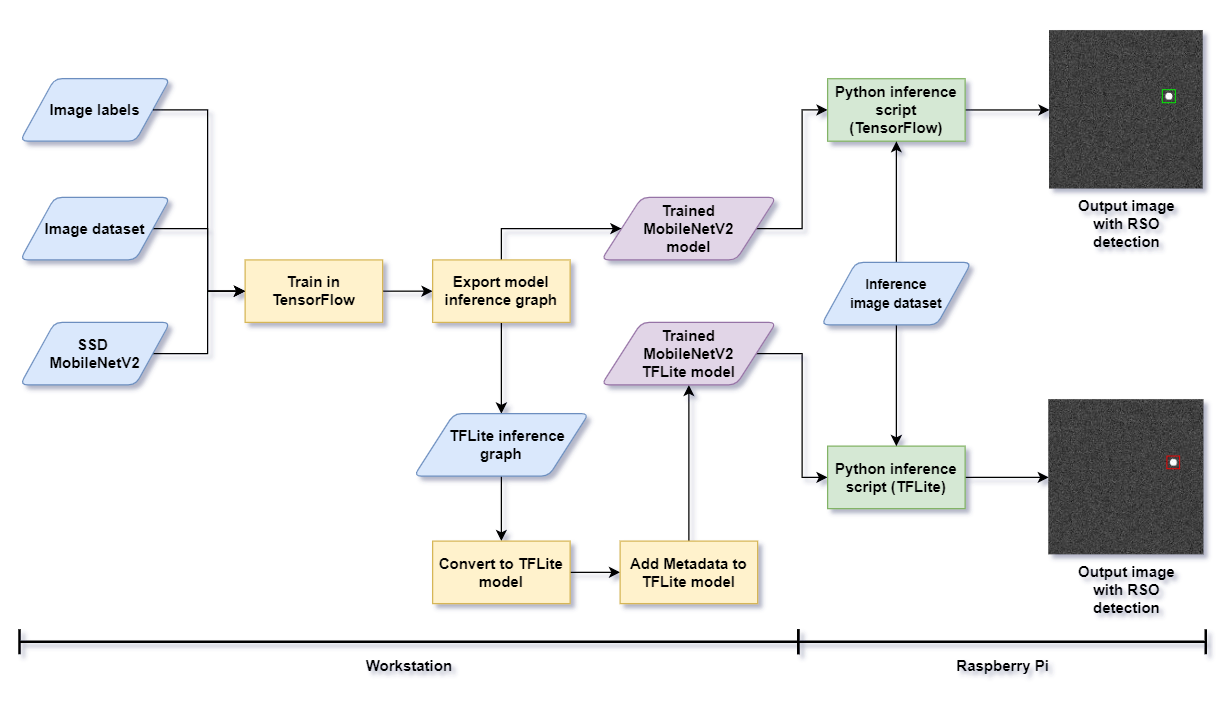}
    \end{minipage}%
    \caption{General procedure for training and image inference}
    \label{fig:Flowchart}
\end{figure}

Beginning on the workstation, a the MobileNetV2 model was trained locally on the custom SatSim dataset via TensorFlow (TF) using the built-in object detection and model libraries. After training, two separate inference graphs were exported in anticipation for performance comparisons between the original model and its TFLite version; however, TFLite models require additional steps before final deployment. Once both versions of the trained model were ready, they were transferred to the Raspberry Pi for inference testing. The general end-to-end outline of this process has been used successfully in other MobileNetV2 applications for object detection as well and is shown in detail in Figure \ref{fig:Flowchart}.\cite{posada2022autonomous,posada2022nn} EfficientDetLite, included as a part of the TFLite Model Maker object detection library, was also trained on the same dataset. Optimized specifically for TFLite implementation, the model does not need to be converted after training, thus serving as a performance comparison for the converted MobileNetV2 model. For this reason, the training procedure for these specific models slightly deviates from that for MobileNetV2.

\section{Performance}
\begin{figure}[!b]
    \centering
        \centering
    \begin{minipage}{0.5\textwidth}
        \centering
        \includegraphics[width=0.92\textwidth]{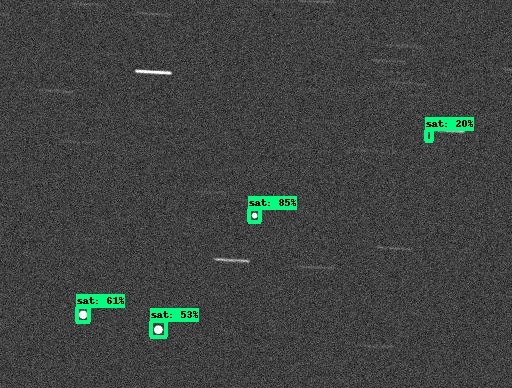}
    \end{minipage}%
    \begin{minipage}{0.5\textwidth}
        \centering
        \includegraphics[width=0.92\textwidth]{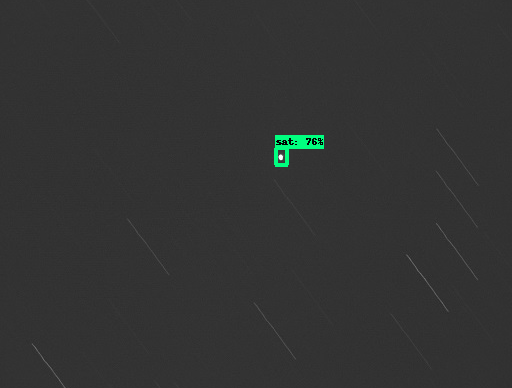}
    \end{minipage}
        \centering
    \begin{minipage}{0.5\textwidth}
        \centering
        \includegraphics[width=0.92\textwidth]{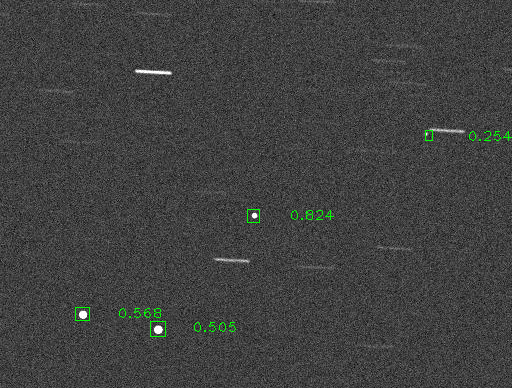}
    \end{minipage}%
    \begin{minipage}{0.5\textwidth}
        \centering
        \includegraphics[width=0.92\textwidth]{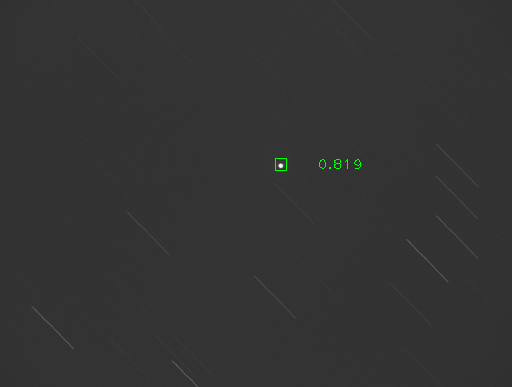}
    \end{minipage}
    \caption{Sample inference results from MobileNetV2 with SatSim generated images (left) and actual satellite imagery from the Celestron RASA telescope (right). Top: TensorFlow. Bottom: TFLite}
    \label{fig:MobileNet}
\end{figure}

\begin{figure}[!htb]
    \centering
        \centering
    \begin{minipage}{0.5\textwidth}
        \centering
        \includegraphics[width=0.95\textwidth]{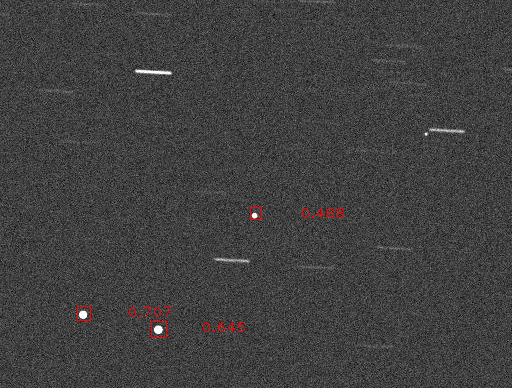}
    \end{minipage}%
    \begin{minipage}{0.5\textwidth}
        \centering
        \includegraphics[width=0.95\textwidth]{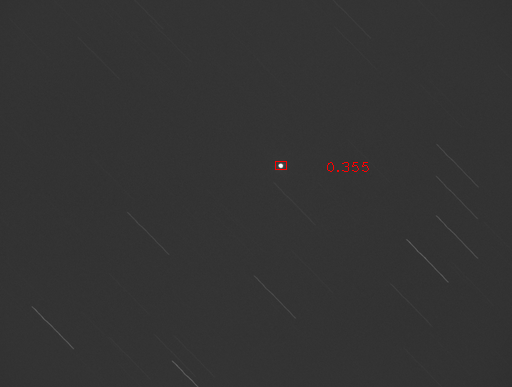}
    \end{minipage}
        \centering
    \begin{minipage}{0.5\textwidth}
        \centering
        \includegraphics[width=0.95\textwidth]{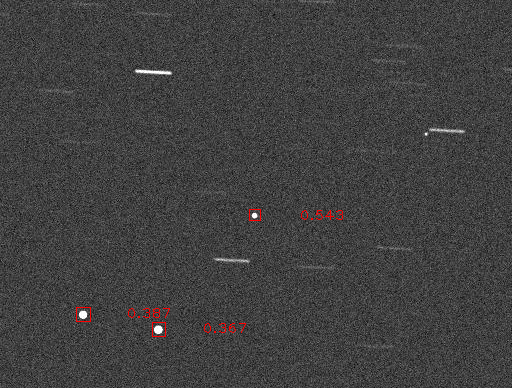}
    \end{minipage}%
    \begin{minipage}{0.5\textwidth}
        \centering
        \includegraphics[width=0.95\textwidth]{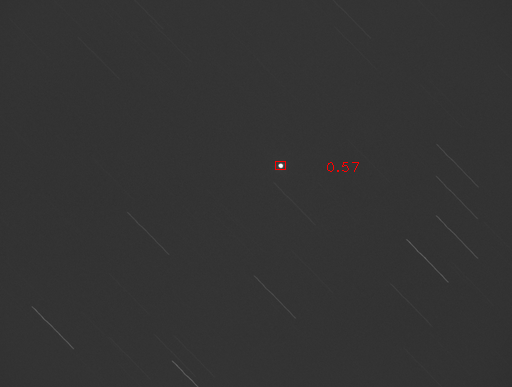}
    \end{minipage}
    \caption{Sample inference results from EfficientDetLite. Top: EfficientDetLite0. Bottom: EfficientDetLite4}
    \label{fig:EffDet}
\end{figure}

As the mobile deployment of autonomous satellite tracking is a significant focus for this work, the minimization of model inference times on the Raspberry Pi was a crucial factor. Keeping this in mind, conversion to TFLite allows for the limited resource availability while keeping inference latency low. The scenarios for tracking satellites (either as small streaks or as point-source dots) were tested with MobileNetV2 and the two versions of EfficientDetLite as well. These models, while trained solely on the SatSim-generated images, were given portions of the real \emph{and} synthetic satellite image databases for inference testing and verification. Results from the MobileNetV2 model are shown in Figure \ref{fig:MobileNet} where confidence levels and RSO detections between the TFlite and TF versions are paired with mixed imagery. For comparison, the inference performance from the EfficientDetLite models can be seen in Figure \ref{fig:EffDet}.

It is worth noting that inference on the two tracking modes requires separate models. This is due to the nature of the target bodies shape profiles completely switching places when changing tracking reference. In early testing, tracking methods where both targets were streaks were investigated; however, issues arose as the only difference is the skewed satellite travel vector compared to the star field. Since the model trained in this paper is looking for specific shapes in the images, accurate detections could not be made in this mode. As a result, only the two modes discussed previously were pursued. An example of this motion can be seen earlier among the labeled images in Figure \ref{fig:dataset}.

\begin{figure}[!h]
    \centering
        \centering
    \begin{minipage}{0.24\textwidth}
        \centering
        \includegraphics[width=0.98\textwidth]{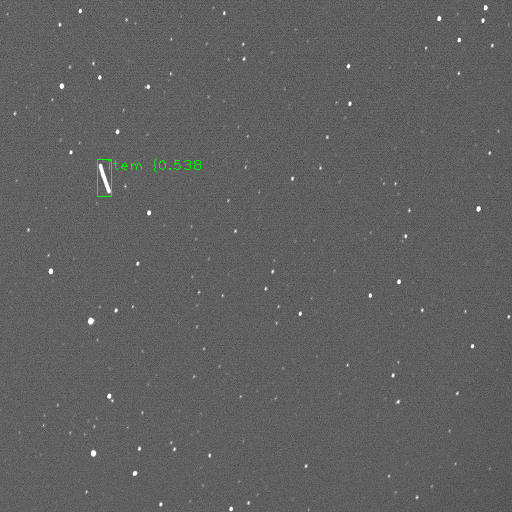}
    \end{minipage}%
    \begin{minipage}{0.24\textwidth}
        \centering
        \includegraphics[width=0.98\textwidth]{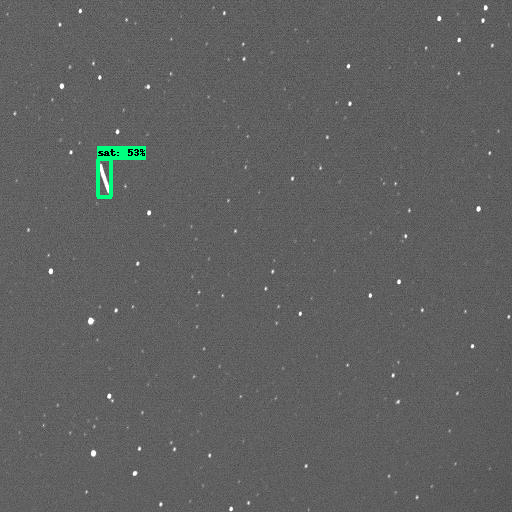}
    \end{minipage}
        \centering
    \begin{minipage}{0.24\textwidth}
        \centering
        \includegraphics[width=0.98\textwidth]{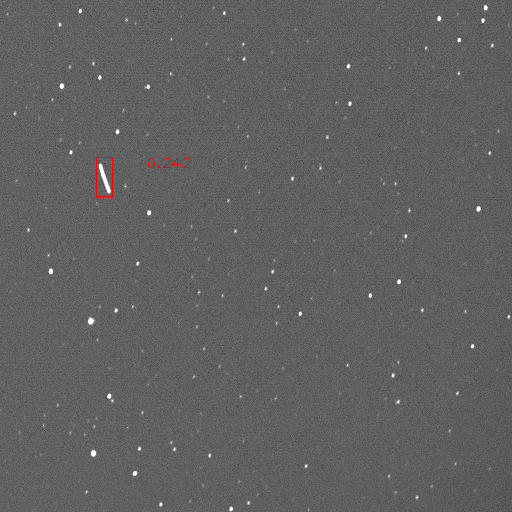}
    \end{minipage}%
    \begin{minipage}{0.24\textwidth}
        \centering
        \includegraphics[width=0.98\textwidth]{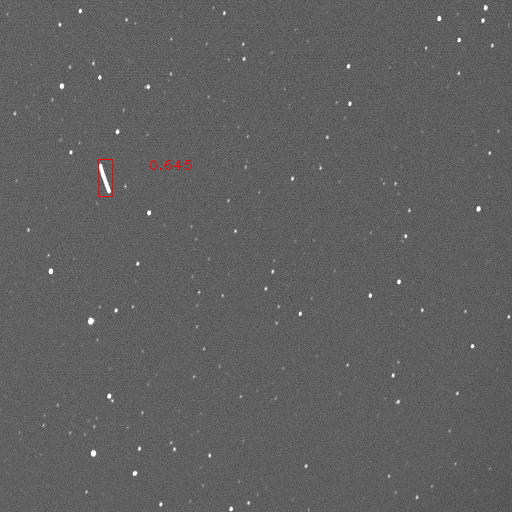}
    \end{minipage}
    \caption{Additional model output from real imagery from opposite tracking mode where RSOs are streaks}
    \label{fig:Streaks}
\end{figure}

While EfficientDetLite0 benefited from the lowest inference times, the confidence levels during object detection suffered a loss in accuracy in many cases. This concept held constant when looking at EfficientDetLite4 where there is a generally higher detection confidence at the cost of significantly longer inference times. The relationship between these two models was consistent across the the majority of the inference images, with only a few outlying cases where the models were close or reversed in detection performance. In situations where RSOs were near stars or were relatively faint in comparison to the other objects in the image, the EfficientDetLite models had the most difficulty. This is clear in Figures \ref{fig:MobileNet} \& \ref{fig:EffDet} where one of the RSOs is completely missed due to the proximity to the streaking star. The difficulty was also reflected in the MobileNetV2 results as, although detected, the confidence was at or barely above the imposed 20\% threshold during testing. Similar challenges were present when the reference was switched as well since the brightness or single stars or star clusters can be difficult to distinguish from RSOs.

\begin{table}[]
\small
    \caption{Average inference times over 100 varied images}
\begin{tabular}{|c|c|c|c|c|}
\hline
\begin{tabular}[c]{@{}c@{}}Object Detection \\ Model\end{tabular}        & SSD-MobileNetV2 & \begin{tabular}[c]{@{}c@{}}SSD-MobileNetV2\\ (TFLite Converted)\end{tabular} & EfficientDetLite0 & EfficientDetLite4 \\ \hline
\begin{tabular}[c]{@{}c@{}}Average Inference\\ Time {[}$s${]}\end{tabular} & 0.992           & 0.649                                                                        & 0.121             & 1.24              \\ \hline
\end{tabular}
    \label{tab:inference}
\end{table}

\begin{figure}[!htb]
    \centering
        \centering
    \begin{minipage}{.49\textwidth}
        \centering
        \includegraphics[width=0.9\textwidth]{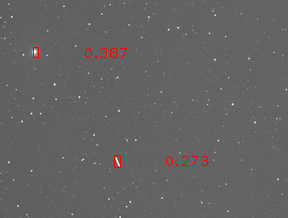}
    \end{minipage}
    \begin{minipage}{.49\textwidth}
        \centering
        \includegraphics[width=0.9\textwidth]{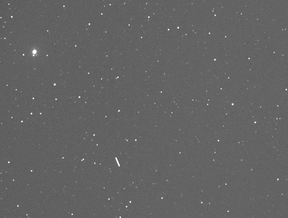}
    \end{minipage}%
    \caption{False positive due to star clusters. Left: Inference detection. Right: Original image}
    \label{fig:difficulty}
\end{figure}

\section{Results}
The result of this work is a CNN model that is be able to correctly detect satellites in star imagery (both simulated and real), while trained on a limited dataset. The performance of the model was checked by running inference on an external database of real satellite imagery and comparing the computation time and detection confidence. The detected locations of satellites in imagery can then be used as measurements for orbit determination. Having an effective model based synthetic imagery presents a number of benefits, primarily in the ease training data acquisition. Observation time and resources which would have otherwise been used for gathering training images could be used to enhance or extend final model deployment instead. Additionally, this method of satellite detection is ideal for embedded systems and mobile implementation where computational resources needed for inference can be limited. Note that the results in Table \ref{tab:finalresults} do not take confidence values of each observation above the threshold into account for model comparison.

\begin{table}[]
\small
    \caption{Model performance with 50 inference targets}
\begin{tabular}{|c|c|c|c|c|}
\hline
\begin{tabular}[c]{@{}c@{}}Object Detection \\ Model\end{tabular} & SSD-MobileNetV2 & \begin{tabular}[c]{@{}c@{}}SSD-MobileNetV2\\ (TFLite Converted)\end{tabular} & EfficientDetLite0 & EfficientDetLite4 \\ \hline
Precision                                                         & 0.9574          & 0.9783                                                                       & 0.9767            & 0.9778            \\ \hline
Recall                                                            & 0.9783          & 1.0                                                                          & 0.913             & 0.9565            \\ \hline
\begin{tabular}[c]{@{}c@{}}F1 \\ (threshold = 0.25)\end{tabular}  & 0.9677          & 0.989                                                                        & 0.9438            & 0.967             \\ \hline
\end{tabular}

    \label{tab:finalresults}
\end{table}

\section{Conclusions and Future Work}
Based on this work, it has been shown that synthetic imagery can be used in practical object detection applications for tracking satellites. Due to the robust performance of the model on a small device, a more advanced implementation of this concept could be pursued where RSOs are tracked in real-time from mobile ground-based telescope observations. Future research can be done to optimize the detection of RSOs in non-ideal conditions where the objects are relatively dim or close in proximity to stars in the images. Further augmentations could also provide accurate automated orbit determination informed by satellite detection models.

\section*{Acknowledgments}
This work was partially supported by the National Defense Science and Engineering Graduate (NDSEG) Fellowship Program. 

\clearpage
\bibliographystyle{AAS_publication}   
\bibliography{sat_ML}   

\end{document}